# ProMamba: Prompt-Mamba for polyp segmentation

Jianhao Xie[1], Ruofan Liao*[1,2], Ziang Zhang[1], Sida Yi[1], Yuesheng Zhu[1], Guibo Luo[1]

[1] School of Electronic and Computer Engineering, Peking University, China
[2] Aberdeen Institute of Data Science and Artificial Intelligence, South China Normal University
zhuys@pku.edu.cn, luogb@pku.edu.cn

**Abstract.** Detecting polyps through colonoscopy is an important task in medical image segmentation, which provides significant assistance and reference value for clinical surgery. However, accurate segmentation of polyps is a challenging task due to two main reasons. Firstly, polyps exhibit various shapes and colors. Secondly, the boundaries between polyps and their normal surroundings are often unclear. Additionally, significant differences between different datasets lead to limited generalization capabilities of existing methods. To address these issues, we propose a segmentation model based on Prompt-Mamba, which incorporates the latest Vision-Mamba and prompt technologies. Compared to previous models trained on the same dataset, our model not only maintains high segmentation accuracy on the validation part of the same dataset but also demonstrates superior accuracy on unseen datasets, exhibiting excellent generalization capabilities. Notably, we are the first to apply the Vision-Mamba architecture to polyp segmentation and the first to utilize prompt technology in a polyp segmentation model. Our model efficiently accomplishes segmentation tasks, surpassing previous state-of-the-art methods by an average of 5% across six datasets. Furthermore, we have developed multiple versions of our model with scaled parameter counts, achieving better performance than previous models even with fewer parameters. Our code and trained weights will be released soon.

**Keywords:** Medical image segmentation, Vision State Space Models, Prompt segmentation.

## 1. Introduction

In the current medical field, the treatment of cancer has always been a challenging problem to solve, and cancer prevention is extremely important and meaningful. Polyps are abnormal proliferative tissues on the mucosal surface that can appear in various parts of the human body, such as the colon, rectum, stomach, and nasal cavity, and are closely related to cancer, as many polyps have the potential to develop into cancerous lesions. Therefore, the task of segmenting polyps during colonoscopy is of great significance for the prevention of intestinal cancer and the implementation of clinical surgeries. Polyp segmentation is a challenging task in the field of medical image processing, involving the accurate extraction and separation of polyp regions

---

*work as an intern at the School of Electronic and Computer Engineering, Peking University

from medical images. The main purpose of polyp segmentation is to assist doctors in diagnosing, treating, and assessing polyps.

However, polyp segmentation has always been a hard task in medical image segmentation due to several reasons. Firstly, polyps vary greatly in size, resulting in significant volumetric differences. Secondly, polyps exhibit considerable differences in shape and color. For example, some polyps have lobulated or papillary shapes, making it difficult to determine their boundaries during segmentation, which can lead to incomplete or over-segmentation. Thirdly, polyps are often difficult to separate completely from surrounding normal tissues, resulting in irregular and blurred edges, further complicating accurate segmentation.

In recent years, deep learning has demonstrated impressive performance in polyp segmentation. The introduction of the U-Net [1] model has shown its excellent performance in medical image segmentation, including polyp segmentation tasks, significantly improving segmentation accuracy. U-Net++ [2], as an improvement on the U-Net structure, enhances the efficiency of cross-layer connections and achieves good results. Subsequent models like ResUNet [3] and ResUNet++ [4] incorporate residual connections into the U-Net structure, mitigating gradient vanishing issues and improving segmentation effectiveness. PraNet [5] introduces a reverse attention module that establishes relationships between target regions and boundaries, leveraging the intrinsic relationship between edges and regions to further enhance segmentation accuracy. TGANet [6] utilizes finer-grained text to provide cues for the specialized handling of polyps in different states, achieving promising results. Additionally, the emergence of the Segment Anything Model (SAM) [7] demonstrates the significant assistance of prompts in segmentation tasks, validating the effectiveness of prompt-based segmentation.

But the methods mentioned above have certain limitations. They often exhibit low segmentation accuracy for polyps with irregular shapes or small samples. Furthermore, these models have limited generalization capabilities, making it challenging to achieve good segmentation results on unseen datasets. They are generally suitable only for specific tasks. While the pre-training dataset for the SAM model primarily consists of natural images, there is a domain gap compared to images used for polyp segmentation, resulting in subpar performance in medical image segmentation. Nevertheless, we are interested in the prompt-based segmentation in SAM, which exhibits strong generalization capabilities, and aim to apply such a technique to our model.

Concurrently, we are aware of a novel architecture in the NLP domain called Mamba [14], which surpasses the performance of Transformer while significantly reducing computational requirements. There have been efforts to migrate Mamba from the NLP domain to the CV domain, with vision-Mamba [15] demonstrating its excellent performance in computer vision. The Mamba-UNet [19] is the first work to migrate Mamba to medical image segmentation. It modularizes the transformer module of Swin-UNet [22] into Mamba. Mamba-UNet has been achieved excellent results on several datasets, surpassing Swin-UNet and U-Net. The VM-UNET [20] shares similar ideas with Mamba-UNet, but it focuses on skin cancer detection and has achieved good results on the ISIC2017 [23] and ISIC2018 [24] datasets. Meanwhile, the subsequent VM-UNET-V2 [21] rethought the architecture of Vision Mamba UNet and proposed an attention-based deep supervision module, which improved segmentation accuracy and reduced model parameters. Simultaneously, U-Mamba [25] has demonstrated exceptional performance in 3D medical image segmentation tasks, surpassing previous sota methods. However, the existing methods simply replace the transformer with Mamba, and there is still considerable room for optimization in the model architecture.

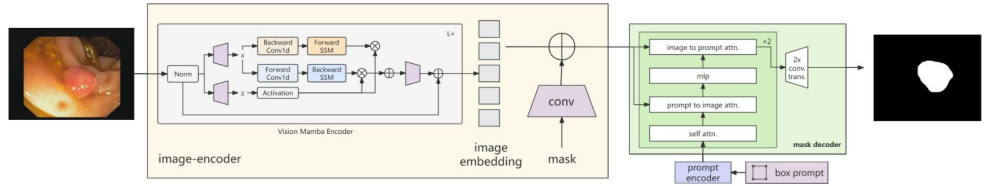

**Fig. 1.** Detailed structure of the model. The model consists of three parts: image-encoder, prompt-encoder and mask-decoder. It is worth noting that the addition of masks in the figure only exists during the training phase.

Therefore, we introduce a polyp segmentation model based on Prompt-Mamba. This model utilizes vision-mamba as its backbone and incorporates prompt assistance, exhibiting excellent generalization performance on downstream datasets, surpassing current methods.

Our contributions can be summarized as follows:
1. We are the first to apply Vision-Mamba (ViM) technology to polyp segmentation, exploring the potential of ViM technology in medical image segmentation.
2. We propose a concise and effective model structure that incorporates prompt technology, resulting in outstanding generalization capabilities.
3. We scale the model differently in experiments and achieve good results even with fewer parameters compared to previous methods.

## 2. Method

### 2.1 Network Structure

The model in this paper is divided into three parts, including a lightweight Image-encoder, which is composed of vision-Mamba, Prompt-encoder, and Mask-decoder. The specific schematic is shown in Fig 1.

**Image-encoder**: The specific parameters include the patch-size of 16, the image-embedding of 768, and 24 layers of basic vision-Mamba. Each basic vision-Mamba layer comprises a fundamental Mamba module. Given the uniqueness of vision tasks, in addition to the standard forward branch, a backward SSM branch is also included, resembling a bidirectional sequence that better captures image feature values. The input to the image-encoder is a medical image, and the final output is the image-embedding. Furthermore, during the training phase, we introduce the mask into the model to further enrich semantic information. The combination of the image-embedding processed by the image-encoder and the mask treated through convolution forms our genuine image-embedding.

Table 1. Comparison of result data.

| Method | Parameter | | CVC-300 | CVC-ClinicDB | CVC-ColonDB | ETIS-LaribPolypDB | Kvasir | BKAI | Mean |
|---|---|---|---|---|---|---|---|---|---|
| U-Net | 4M | Dice | 0.8298 | **0.9073** | 0.6370 | 0.4933 | 0.8446 | 0.7394 | 0.7419 |
|  |  | IoU | 0.7415 | 0.8472 | 0.5384 | 0.4079 | 0.7671 | 0.6502 | 0.6587 |
| PraNet | 32M | Dice | 0.8722 | 0.8996 | 0.7061 | 0.6179 | 0.8897 | 0.7945 | 0.7966 |
|  |  | IoU | 0.7966 | 0.8396 | 0.6362 | 0.5586 | 0.8179 | 0.7111 | 0.7266 |
| TGANet | 19M | Dice | 0.8679 | 0.8935 | 0.7144 | 0.5949 | 0.8912 | 0.7587 | 0.7867 |
|  |  | IoU | 0.7936 | 0.8390 | 0.6300 | 0.5163 | 0.8204 | 0.6829 | 0.7137 |
| U-Net++ | 47M | Dice | 0.8525 | 0.8821 | 0.6529 | 0.4647 | 0.8453 | 0.7342 | 0.7383 |
|  |  | IoU | 0.7602 | 0.8092 | 0.5596 | 0.3894 | 0.7741 | 0.6504 | 0.6570 |
| nnU-Net | 46M | Dice | 0.8779 | 0.9014 | 0.7682 | 0.5997 | 0.9053 | 0.8385 | 0.8151 |
|  |  | Iou | 0.8023 | **0.8573** | 0.6971 | 0.5281 | 0.8542 | 0.7711 | 0.7516 |
| SAM-b | 91M | Dice | 0.7101 | 0.5267 | 0.535 | 0.4821 | 0.6710 | 0.6401 | 0.5941 |
|  |  | IoU | 0.6397 | 0.4447 | 0.4595 | 0.4032 | 0.5798 | 0.5611 | 0.5146 |
| SAM-l | 309M | Dice | 0.7251 | 0.5747 | 0.5129 | 0.5839 | 0.7877 | 0.6969 | 0.6468 |
|  |  | IoU | 0.6641 | 0.4985 | 0.4461 | 0.5187 | 0.7030 | 0.5289 | 0.5598 |
| SAM-h | 637M | Dice | 0.6513 | 0.6007 | 0.5116 | 0.6395 | 0.6934 | 0.6805 | 0.6295 |
|  |  | IoU | 0.5891 | 0.5310 | 0.4446 | 0.5687 | 0.6118 | 0.6160 | 0.5602 |
| Ours | 102M | Dice | **0.8909** | 0.8879 | **0.8202** | **0.7713** | **0.8863** | **0.8603** | **0.8528** |
|  |  | IoU | **0.8075** | 0.8140 | **0.7124** | **0.6630** | **0.8078** | **0.7724** | **0.7628** |

**Prompt-encoder:** In the comparison between point-prompt and box-prompt, we adopt box-prompt to carry out the prompting task. This is because, during the process of training, box-prompt exhibits greater stability compared to point-prompt, which is conducive to training a more stable model. Box-prompt is a type of prompt that mimics human user interaction. In the proposed prompt-encoder, we generate box-prompt by offsetting the randomly original mask. During the training phase, multiple prompts can be generated, and the final result is obtained by using the average value. During the training process, our operation of boxes is similar to that of SAM, where we record the positional coordinates of the four corners of the box and then convert these coordinates into an embedding format. The input is the coordinates of the box, and the output is the prompt-embedding.

**Mask-decoder**: In the design process of our mask-decoder, we drew inspiration from the excellent mask-encoder design for SAM, striving to ensure the lightness of the mask-decoder. We adopted a similar design approach. Our decoder design is shown in Fig 1. Our inputs are the image-embedding and prompt-embedding. After entering the prompt-embedding, it first goes through a layer of self-attention, and then undergoes two layers of cross-attention with the image-embedding. Finally, it is enlarged to its normal size through convolution, and the output is the mask.

## 2.2     Loss function

Our loss function is chosen as a linear combination of Focalloss [17], Diceloss. Focalloss is an improved version of CELoss that is better able to handle unbalanced data, which is why we chose it.The reason for choosig Diceloss is that dice is important criteria for measuring segmentation criteria in image segmentation, so we use these two losses to improve segmentation quality.

$$\mathcal{L}_{loss} = \mathcal{L}_{Diceloss} + \alpha * \mathcal{L}_{Focalloss} \quad (1)$$

In the above equation, α is hyperparameters that represent the weights of different losses.

## 3.     Experiments

We followed the classical experimental design by training on a merged dataset combining two datasets and conducting tests separately on the test sets of these two datasets. Subsequently, we migrated our model to four unseen datasets for further testing. We conducted certain parameter scaling on our model, proposing multiple versions suitable for various scenarios. Additionally, we performed ablation experiments to explore the key factors that affect the model's performance. Detailed

Table 2. Comparison of scaling results for model parameter quantity.The parameters for Ours are depth 24 and emb 768.

| Model | Parameter | | CVC-300 | CVC-ClinicDB | CVC-ColonDB | ETIS-LaribPolypDB | Kvasir | BKAI | Mean |
|---|---|---|---|---|---|---|---|---|---|
| Emb192 | 11M | Dice | 0.8663 | 0.8335 | 0.7903 | 0.7403 | 0.8613 | 0.8425 | 0.8223 |
|  |  | IoU | 0.7688 | 0.7364 | 0.6710 | 0.6244 | 0.7711 | 0.7478 | 0.7199 |
| Emb384 | 30M | Dice | 0.8894 | 0.866 | 0.8169 | 0.7612 | 0.8777 | 0.8561 | 0.8445 |
|  |  | IoU | 0.8058 | 0.778 | 0.7081 | 0.6454 | 0.7963 | 0.7654 | 0.7498 |
| Depth12 | 54M | Dice | 0.8563 | 0.8536 | 0.8082 | 0.7512 | 0.8802 | 0.8547 | 0.8340 |
|  |  | IoU | 0.7613 | 0.7747 | 0.6957 | 0.6400 | 0.8033 | 0.7664 | 0.7402 |
| Depth18 | 78M | Dice | 0.8660 | 0.8725 | **0.8204** | **0.7717** | 0.8856 | 0.8563 | 0.8454 |
|  |  | IoU | 0.7674 | 0.7865 | 0.7097 | 0.6587 | 0.8077 | 0.7628 | 0.7488 |
| Ours | 102M | Dice | **0.8909** | **0.8879** | 0.8202 | 0.7713 | **0.8863** | **0.8603** | **0.8528** |
|  |  | IoU | **0.8075** | **0.8140** | **0.7124** | **0.6630** | **0.8078** | **0.7724** | **0.7628** |

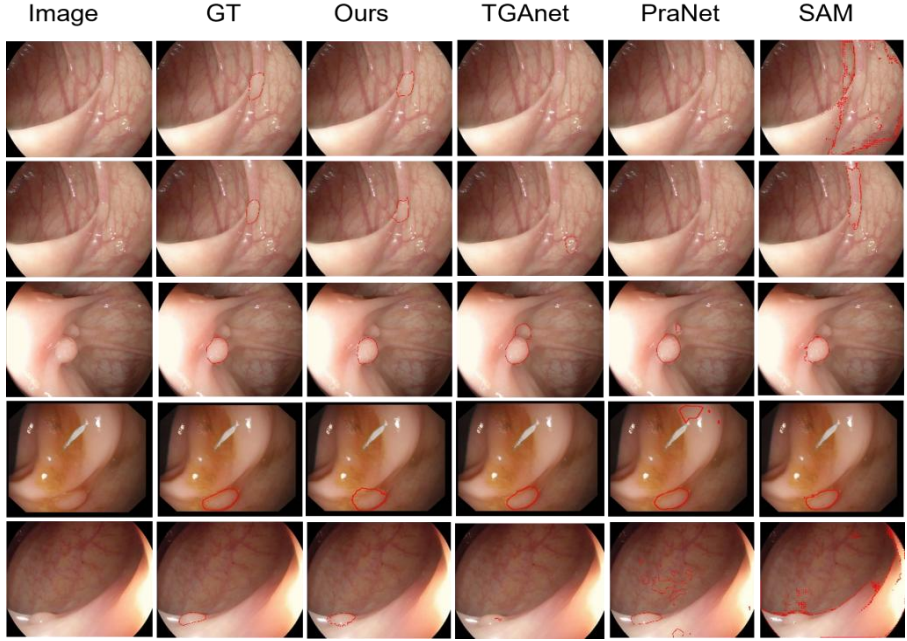

**Fig. 2** Segmentation result image. In the method diagram, we overlay the mask generated by the model with the image, using red lines to indicate it. The SAM mentioned here refers to the SAM-b type.

information will be presented in the following chapters.

### 3.1    Datasets and Baselines

Our experiment consists of six datasets, ETIS [8], CVC-ClinicDB/CVC-612 [9], CVC-ColonDB [10], EndoScene-CVC-300 [11], Kvasir [12] and BKAI [13]. All six data sets are public. For the data set partition, we followed the pranet design except for the BKAI data set, and the training set consisted of 80% of Kvasir and CVC-ClinicDB, the validation set consisted of 10% of Kvasir and CVC-ClinicDB, and the test set consisted of the rest of the data sets. At the same time, we added BKAI to the test dataset.

About baseline, We compare our method with five SOTA medical image segmentation methods: U-Net, U-Net++, ParNet, TGANet and nnU-Net [16]. We also compare the SAM based on prompt segmentation. In the comparison method, we try to completely restore the previous paper methods except SAM, so we keep all the training settings of previous papers, including image size, learning rate, optimizer, etc. For SAM, we chose 3 versions of SAM model, including SAM-b, SAM-l, SAM-h, and directly use its original pre-trained model for downstream task inference.

Our basic experimental environment is python 3.10, torch 2.1.1 +cu118. We use batch-size of 16, optimizer Adam, initial learning rate of 1e-4, no decay rate, loss function of, focalloss combined with diceloss, total epochs of 200. Our input-size is 512 x 512. The final criteria are Dice and IoU. At the same time, our method does basic data enhancement, including horizontal flip, vertical flip, image rotation, random masking and so on.

### 3.2 Experimental Results

Our results are divided into two main parts. One is the comparison of results on validation part of kvasir and CVC-ClinicDB datasets, and the other is the comparison of results on several other unseen datasets such as ETIS, CVC-ColonDB, EndoScene-CVC300, BKAI, etc.

For inference stage, we uniformly choose the best model rather than the last model. At the same time, we display the relevant parameters of the model in the Table 2 to avoid the situation that the performance is better because the model parameters are too large. As can be seen from Table 1 and Table 2, the performance of our model is slightly lower than that of the best method on the validation of the dataset that has been seen. The reason may be that the number of images for training is too small. However, in comparison with the results of the unseen datasets, our results are 5% better than the performance of the previous best methods. The reasons for this results are, firstly, because our advanced backbone uses the excellent feature extraction ability of vision-Mamba, and secondly, the box-prompt technology we use is very versatile. Prompt can be applied for various kind of data, and can greatly enhance the generalization of the model in unseen datasets. It is worth noting that in Table 2, our smallest model with only 11M parameters is only one-third of the parameters of ParNet and one-fourth of nnU-Net. However, the performance of our model still surpasses previous methods, strongly demonstrating the excellence of our model architecture.

**Table 3.** Comparison of ablation experiment results. Among them, "Ours" refers to the combination of backbone, backward_SSM, and input_mask.

| Mthod | | CVC-300 | CVC-ClinicDB | CVC-ColonDB | ETIS-LaribPolypDB | Kvasir | BKAI | Mean |
|---|---|---|---|---|---|---|---|---|
| Backbone | Dice | 0.8830 | 0.8617 | 0.8091 | 0.7591 | 0.8766 | 0.8539 | 0.8405 |
| | IoU | 0.7952 | 0.7769 | 0.7005 | 0.6546 | 0.8007 | 0.7624 | 0.7483 |
| Backbone + backward_SSM | Dice | 0.8739 | 0.8690 | 0.8150 | 0.7648 | 0.8862 | 0.8561 | 0.8441 |
| | IoU | 0.7815 | 0.7814 | 0.7080 | 0.6545 | 0.8012 | 0.7675 | 0.7490 |
| Backbone + input_mask | Dice | 0.8864 | 0.8703 | 0.8174 | **0.7741** | 0.8801 | 0.8548 | 0.8471 |
| | IoU | **0.8091** | 0.7904 | 07113 | 0.6628 | 08058 | 0.7680 | 0.7579 |
| Ours | Dice | **0.8909** | **0.8879** | **0.8202** | 0.7713 | **0.8863** | **0.8603** | **0.8528** |
| | IoU | 0.8075 | **0.8140** | **0.7124** | **0.6630** | **0.8078** | **0.7724** | **0.7628** |

### 3.4    Ablation Study

The ablation experiment we conducted mainly includes two parts, one is the change of model component parameters, including vision-Mamba layers in image-encoder and embedding length. The second is ablation of model components, including the component about backward SSM and the component about add the mask during training .

To ensure the reliability of the contrast effect during parameter changes and model components ablation, we only change one parameter or model component at a time while keeping other parameters and model components unchanged.

**Parameter changes:** The vision-Mamba layers are set to 24, 18 and 12 for contrasts, and the embedding lengths are set to 768, 384 and 192 for contrasts. Our specific results in the Table 2 reflects the varying trends of model performance indicators as the number of model parameters increases or decreases. By comparing the Dice coefficient and IoU values under different parameter quantities, we can gain a deeper understanding of the relationship between model performance and the number of parameters, providing a solid basis for model optimization and selection. The performance of the model is positively correlated with vision-Mamba layers and embedding length. As the number of layers and embedding length increases, the effectiveness of the model is enhanced.

**Component ablation:** We use the largest model(102M) for ablation experiments. As can be seen from the Table 3, ablation of these two components backward_SSM and input_mask does reduce the performance of the model. As can be seen from Table 3, when these two components are absent, the performance of the model exhibits a certain degree of decline. We prove the validity of our backward SSM and input-mask additions in training phase.

## 4.    Conclusion

In this paper, we propose a segmentation model based on prompt-Mamba, which is the first one to incorporate prompt and vision-Mamba techniques into polyp segmentation. Vision-Mamba technology endows the model with strong ability to extract image features, while box-prompt brings good generalization to the model. This enables the model to achieve good results in several unseen tasks, and the effect greatly exceeds that of previous models. In the future, we'll explore more kinds of prompts, similar to text prompts, and collect more data to develop pre-trained models for improve results and generalization.